\documentclass[twoside]{article}
\usepackage{proceed2e}
\usepackage{times}

\usepackage{algorithm}
\usepackage{algorithmic}
\usepackage{latexsym}
\usepackage{amsmath}
\usepackage{amssymb}
\usepackage{graphicx}
\usepackage{braket}
\usepackage{color}
\usepackage{natbib}

    {\mbox{}\hfill\mbox{$\blacksquare$}\\%
}

\definecolor{darkgreen}{rgb}{0.0,0.5,0.0}
\definecolor{darkyel}{rgb}{0.5,0.5,0.0}

\newcommand{\comment}[1]{\colorbox{yellow}{#1}}
\renewcommand{\comment}[1]{}

\newcommand{\markercolor}[2]{{#2}}

\newcommand{\marker}[1]{\markercolor{red}{#1}}
\newcommand{\markerb}[1]{\markercolor{blue}{#1}}
\newcommand{\markerc}[1]{\markercolor{darkgreen}{#1}}
\newcommand{\markerd}[1]{\markercolor{magenta}{#1}}
\newcommand{\markere}[1]{\markercolor{magenta}{#1}}
\newcommand{\markerf}[1]{\markercolor{darkyel}{#1}}
\newcommand{\markerg}[1]{\markercolor{red}{#1}}

\newcommand{\jfootnote}[1]{} 

\let\mathbf=\boldsymbol
\let\cite=\citep
\def\x{x}

\usepackage[dvipdfm]{hyperref}

\newcommand\DOWNTO{ \textbf{downto} }
\newcommand\ttt{{\mathit{i}}}

\title{
Restricted Collapsed Draw: 
Accurate Sampling for \\ 
Hierarchical Chinese Restaurant Process Hidden Markov Models
}
\author{}

\begin{document} 

\setlength{\textfloatsep}{15pt plus30pt}
\setlength{\abovedisplayskip}{5pt plus2pt minus.5pt}
\setlength{\belowdisplayskip}{5pt plus2pt minus.5pt}

\maketitle
\begin{abstract} 
\marker{
We propose a restricted collapsed draw (RCD) sampler, a general Markov chain Monte Carlo sampler 
of simultaneous draws from \markerg{a} hierarchical Chinese restaurant process (HCRP) with restriction. 
Models that require simultaneous draws from \markerg{a} hierarchical Dirichlet process with restriction, 
such as infinite Hidden markov models (iHMM), were difficult 
to enjoy benefits of \markerg{the} HCRP due to combinatorial explosion in calculating distributions of coupled draws.
By constructing a proposal of seating arrangements (partitioning) and stochastically accepts the proposal by
the Metropolis-Hastings algorithm, the RCD sampler makes accurate sampling for complex combination of draws while
retaining efficiency of HCRP representation.  Based on the
RCD sampler, we developed a series of sophisticated sampling algorithms for iHMM\markerg{s}, including blocked Gibbs sampling, beam sampling, 
and split-merge sampling, that outperformed conventional iHMM samplers in experiments}.
\vspace{-2ex}
\end{abstract} 

\section{Introduction}

\begin{figure}
\centering{\includegraphics{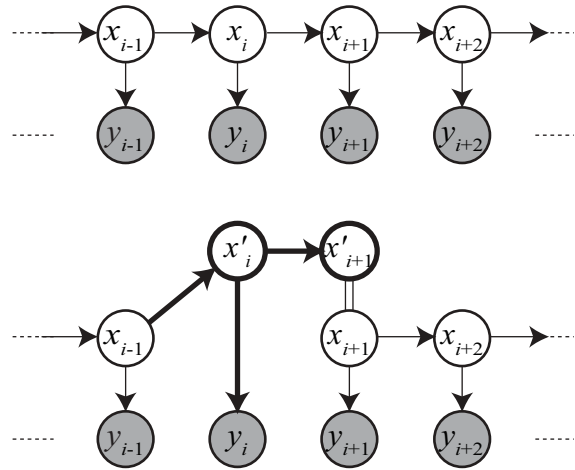}}
\caption{
Step-wise Gibbs sampling in iHMM\@. 
Since the Dirichlet process prior is posed on \textit{transitions} in iHMM,  resampling $\x_i$ involves taking
 two transitions, $\x_{i-1} \rightarrow \x_i$ and $\x_i \rightarrow \x_{i+1}$, simultaneously. 
 In this case, we consider distribution of two draws $(x'_i, x'_{i+1})$ with restriction that 
 the draws are consistent with remaining sequence, i.e., $x'_{i+1} = x_{i+1}$.
}\label{fig:resamp}
\end{figure}

Existing sampling algorithms for infinite hidden Markov models (iHMMs,
also known as the hierarchical Dirichlet process HMMs) \cite{beal2002infinite,teh2006hierarchical} 
do not use a hierarchical Chinese restaurant process (HCRP) \cite{teh2006hierarchical}, 
which is a way of representing the predictive distribution of a hierarchical Dirichlet process (HDP) by 
collapsing\markerd{, i.e. integrating out, the \markere{underlying} distribution}\markerc{s} \markere{of} the \markerc{Dirichlet process (DP)}.
\markere{While} \marker{an HCRP representation provides efficient sampling} for many other models based on an HDP \cite{teh2006bayesian,mochihashi2008infinite}
\marker{through reducing the dimension of sampling space}, 
it has been considered rather ``awkward'' \cite{teh2006hierarchical} to use \markerg{an} HCRP for iHMMs, due to the difficulty in handling 
coupling between random variables.  In the simplest case, consider step-wise Gibbs sampling from an iHMM
\markerc{defined as $\mathbf{\pi}_{k} \sim \mathrm{DP}(\mathbf{\beta}, \alpha_0)$ and $\mathbf{\beta} \sim \mathrm{GEM}(\gamma)$}.
 Given $\x_1, \ldots, \x_{i-1}, \x_{i+1}, \ldots, \x_T$,
resampling \marker{hidden state $\x_i$} at time step $i$ actually consists of two draws (Figure~\ref{fig:resamp})%
, \markerb{$x'_i \sim \mathbf{\pi}_{\x_{i-1}}$ and $x'_{i+1} \sim \mathbf{\pi}_{\x'_{i}}$}, under the 
restriction \markerb{$(\x'_i, \x'_{i+1}) \in C$} that these draws are consistent with the following sequence, i.e., 
\markerb{$C=\{ (x'_i, x'_{i+1}) | x'_{i+1} = \x_{i+1}  \}  $}%
. Under the HCRP, the two draws are coupled even if $\x_{i-1} \ne x'_i$, \markerc{because 
distributions 
$\mathbf{\pi}_{\x_{i-1}}$, $\mathbf{\pi}_{x'_i}$ as well as the base measure
$\mathbf{\beta}$} are integrated out in an HCRP\@, and 
coupling complicates sampling from the restricted distribution.

To generalize, the main part of the difficulty is to obtain a sample from a restricted joint distribution of simultaneous draws 
from collapsed distributions, which we call \textit{restricted collapsed draw} (RCD).
Consider resampling $L$ draws \markere{simultaneously}, $\mathbf{x} = (\x_{j_1i_1}, \ldots, \x_{j_Li_L})$, from 
\markere{the respective} restaurants $\mathbf{j} = (j_1, \ldots, j_L)$, when we have a restriction $C$ such that $\mathbf{x} \in C$.  
Step-wise Gibbs sampling from iHMM can be fitted into RCD \markerd{with $L=2$} by allowing restaurant index $j_2$ to be 
dependent on the preceding draw $\x_{j_1i_1}$.

In this paper, we point out that it is not enough to consider the distribution of draws.  Since the HCRP 
introduces an additional set of latent variables \markerg{$\mathbf{s}$} that accounts for the seating arrangements of the restaurants,
we have to compute an exact distribution of \markerg{$\mathbf{s}$} as well, under the restriction.
We want to perform sampling from the following conditional distribution,
\begin{align}
p( \mathbf{x}, \mathbf{s} |  C ) = 
{\frac{1}{Z_C}} \, \mathbb{I}[\, \mathbf{x} \in C \,] \; p( \mathbf{x}, \mathbf{s} ) \quad ,
\label{eq:drawcond}
\end{align}
where $Z_C$ is a normalization constant and $\mathbb{I}$ is the indicator function, whose value is 1 if the condition is true and 0 otherwise.
Although non-restricted probability $p( \mathbf{x}, \mathbf{s} )$ can be easily calculated 
for a given $\mathbf{x}$ and $\mathbf{s}$, 
calculating the normalization constant $Z_C$ 
leads to a combinatorial explosion in terms of $L$\@.

\marker{
To solve this issue, we propose the restricted collapsed draw (RCD) sampler, 
which provides accurate distributions of simultaneous draws and seating arrangements from HCRP\@.
The RCD sampler constructs a proposal of seating arrangements using a given proposal of draws, and the pair of proposals 
are stochastically accepted by the Metropolis-Hastings algorithm \cite{hastings1970monte}.
Since the RCD sampler can handle any combination of restricted collapsed draws simultaneously, 
}%
we \markerg{were able to develop} \marker{a series of sampling method for HCRP-HMM}, including
a blocked collapsed Gibbs sampler, a collapsed beam sampler, and a split-merge sampler for HCRP-HMM.
Through experiments we found that our collapsed samplers outperformed their non-collapsed counterparts.

\section{HCRP representation for iHMM}

\subsection{Infinite HMM}

An infinite hidden Markov model (iHMM) \cite{beal2002infinite,teh2006hierarchical} is defined over the following HDP:
\begin{align}
G_0 \sim {}& \mathrm{DP}(\gamma, H)  &   G_k \sim {}& \mathrm{DP}(\alpha_0, G_0)  \quad , \label{eq:hdp}
\end{align}


To see the relation of this HDP to the transition matrix $\mathbf{\pi}$, consider the explicit representation of parameters:
\begin{align}
G_0 = {}& \sum_{k'=1}^\infty \beta_{k'} \phi_{k'}  &
G_k = {}& \sum_{k'=1}^\infty \pi_{k'k} \phi_{k'}\quad ,
\end{align}
where transition probability \markerc{$\mathbf{\pi}_k$ is given as} $\mathbf{\pi}_k \sim \mathrm{DP}(\alpha_0, \mathbf{\beta})$,  $\mathbf{\beta} \sim \mathrm{GEM}(\gamma)$ 
is the stick-breaking construction of DPs \cite{sethuraman1994constructive},
  and $\phi_k \sim H$. 

\begin{figure*}
\centering{
\includegraphics{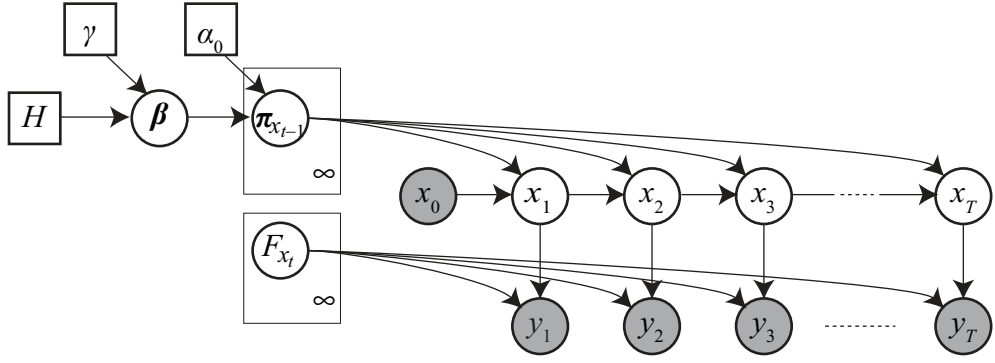}
}
\caption{Graphical Representation of iHMM. }
\label{fig:ihmm}
\end{figure*}

A formal definition for the HDP based on this representation is:
\begin{align}
\mathbf{\beta} | \gamma &{}\sim \mathrm{GEM}( \gamma ) &
\!\!\! \mathbf{\pi}_j | \alpha_0, \mathbf{\beta} &{}\sim \mathrm{DP} ( \alpha_0, \mathbf{\beta} ) \label{eq:hdp2} \\
\!\!\!\!\! \markerb{\x_{ji} | (\mathbf{\pi}_k)_{k=1}^{\infty}} &{}\sim \markerb{\mathbf{\pi}_{j}}
 \quad \phi_k \sim H \!\!\!\!\! &
\markerb{y_{ji} | \x_{ji}} &{}\sim \markerb{F( \phi_{\x_{ji}} ) } \quad ,
\end{align}
\markerb{\markerg{Given an HDP and initial state $x_0$,} we can construct \markere{an infinite} HMM by extracting 
a sequence of draws $x_i$ as $x_i = x_{x_{i-1} i}$, and corresponding
observations $y_i = y_{x_i i}$. }%
Figure~\ref{fig:ihmm} shows a graphical representation of the iHMM\@.

\subsection{HCRP-HMM}

\markerc{%
As another way of representing HDP in iHMM (Eq.~\ref{eq:hdp}), we introduce a hierarchical Chinese restaurant process (HCRP, also known as the Chinese \marker{r}estaurant \marker{f}ranchise), 
which does not need to sample the transition distribution $\mathbf{\pi}$ and its base measure $\mathbf{\beta}$  in Eq.~\eqref{eq:hdp2}:
}\markerb{
\begin{align}
k_{jt} | \gamma &{}\sim \mathrm{CRP}(\gamma)
& 
t_{ji} | \alpha_0 &{}\sim \mathrm{CRP}(\alpha_0) 
\label{eq:alterhcrp}
\\
\x_{j\ttt} &{} = k_{jt_{j\ttt}} \label{eq:alter3}\\
\phi_k &{}\sim H & 
y_{j\ttt} | \x_{j\ttt}, \mathbf{\phi} &{} \sim F( \phi_{\x_{j\ttt}} )  \  \quad .\!\!\!\!\!\!\!\!\!\!\!\!\!\!\!\!\!\!\!\!\!\!\!\!\!\!\!\!\!\!\!\!\!\!\!\!\!\!\!
\label{eq:alter4}
\end{align}%
}%
\markerb{%
Using the Chinese restaurant metaphor, we say that
customer $i$ of restaurant $j$ sits at table $t_{ji}$, which has a dish of an index $k_{jt_{ji}}$.
}

\markerc{
To understand connection between HDP and HCRP, consider a finite model of grouped observations $x_{ji}$,
in which each group $j$ choose a subset of $M$ mixture components from a model-wide set of $K$ mixture components:
\begin{align}
\mathbf{\beta} | \gamma &{}\sim \markerc{\mathrm{Dir}( \gamma/K, \ldots, \gamma/K ) }&
k_{jt} | \mathbf{\beta} &{}\sim \mathbf{\beta}  
\label{eq:alter1}\\
\mathbf{\tau}_j | \alpha_0 &{}\sim \markerc{\mathrm{Dir}( \alpha_0/M, \ldots, \alpha_0/M )}\!\! &
t_{ji} | \mathbf{\tau}_j &{}\sim \mathbf{\tau}_j 
\label{eq:alter2}
\end{align}
\markerc{%
As $K \rightarrow \infty$ and $M \rightarrow \infty$, the limit of this model is HCRP; hence the infinite limit of this model is also HDP\@.
}\markerd{
Equation \eqref{eq:alterhcrp} is derived by taking the infinite limit of K and M after
integrating out $\mathbf{\beta}$ and $\mathbf{\tau}$ in Eqs.~\eqref{eq:alter1} and~\eqref{eq:alter2}.
}
The distribution $\mathbf{\pi}_j$ in Eq.~\ref{eq:hdp2} can be derived from $\mathbf{\tau}_j$ \markerd{and $\mathbf{k}_j$} as follows:
\begin{align}
\nonumber\\[-8mm]
\mathbf{\pi}_j &{}= \sum_t \tau_{jt} \delta_{k_{jt}} \quad .
\end{align}%
}%
%
%
\markerb{%
To consider sampling of $x_{ji}$ using HCRP (Eqs.~\ref{eq:alter3} and~\ref{eq:alter4}), we use count notation} $n_{jtk}$ as the number of customers in restaurant $j$ at table $t$ 
serving the dish of the $k$-th entry, 
and $m_{jk}$ as the number of tables in restaurants the $j$ serving the dish of the $k$-th entry.  We also use dots for marginal 
counts (e.g., $m_{\cdot k}= \sum_j m_{jk}$).  \markerb{Then, we sample table index $t_{ji}$ from the following distribution:
\begin{align}
p(t_{ji} = t | t_{j1}, \ldots, t_{j,i-1}) &{} = \frac{ n_{j t \cdot } }{n_{j \cdot \cdot} \markerc{+ \alpha_0} } \label{eq:hcrp1}\\
p(t_{ji} = t^{new} | t_{j1}, \ldots, t_{j,i-1}) &{} = \frac{ \alpha_0 }{n_{j \cdot \cdot} \markerc{+ \alpha_0} } \quad .
\end{align}
When $t_{ji} = t^{new}$ (i.e., the customer sits at a new table), we need to sample $k_{jt^{new}}$, whose distribution is:
\begin{align}
p(k_{jt} = k | k_{11}, \ldots ) &{} = \frac{ m_{\cdot k} }{m_{\cdot \cdot} \markerc{+ \gamma} } \\
p(k_{jt} = k^{new} | k_{11}, \ldots ) &{} = \frac{ \gamma }{m_{\cdot \cdot} \markerc{+ \gamma} } \quad . \label{eq:hcrp2}
\end{align}
These variables determine the new sample $x_{ji} = k_{jt_{ji}}$.
}
Since $x_{ji}$ does not uniquely determine the state of the HCRP model, we need to
keep latent variables $t_{ji}$ and $k_{jt}$ for subsequent sampling.  We will denote $s^{(j)} = (t_{j1}, t_{j2}, \ldots) $ as the seating arrangement 
in restaurant $j$, $s^{(0)} = (k_{11}, k_{12}, \ldots, k_{21}, \ldots) $ as the seating arrangement in
the root restaurant, and $\mathbf{s}$ as the collection of all seating arrangements, corresponding to the sampled model state.
\jfootnote{ˆÈ'O'Ì review 'Å'Í seating arrangement 'Æ'¢'¤Œ¾—t'à•]"»'ªˆ«'©'Á'½'Ì'¾'¯'ǁA'È'ñ'Æ'¢'¦'΂¢'¢'Ì'©'ȁBn-gram model '̂悤'É
'½ŠK'w'̏ꍇ'Í seating arrangement 'Æ'¢'¤'Ì'ª"IŠm'È‹C'à'·'邪AHMM '̂悤'É2ŠK'w'̏ꍇ'É'Í•Ê'Ì—pŒê'ª' 'é‹C'ª'·'éB'ªA'È'ñ'¾'©'í'©'ç'È'¢EEE}
In Bayesian inference based on sampling, we need a procedure to sample the latent variables,
given the value of new draw $x_{ji}$ and the seating arrangements for other draws $\mathbf{s}$,
which is called as $\mathsf{addCustomer}$.

\markerb{Construction of HCRP-HMM is the same as iHMM, i.e., extracting a sequence of draws $x_i$ given $x_0$ as $x_i = x_{x_{i-1} i}$, and corresponding
observations $y_i = y_{x_i i}$. }



\section{Restricted Collapsed Draw Sampler}

What we want is a sampling algorithm for HCRP-HMM\@. 
As described in the Introduction, the problem can be reduced to an algorithm for sampling 
from $p( \mathbf{x}, \mathbf{s} | C)$, i.e., the distribution of restricted collapsed draw 
with seating arrangements (Eq.~\ref{eq:drawcond}).  

Our idea is to apply the Metropolis-Hastings algorithm \cite{hastings1970monte} 
to the seating arrangements, which stochastically accepts the proposal distribution of 
seating arrangements.  
Although it is hard to directly give proposal distribution $q(\mathbf{s})$ of seating arrangements, 
\markerg{our method constructs $q(\mathbf{s})$ by combining $q_{\mathbf{x}}(\mathbf{x})$ with
$q_{\mathbf{s}}(\mathbf{s}|\mathbf{x})$, another proposal of seating arrangements given the proposed draws,
which is based on the $\mathsf{addCustomer}$ procedure} that is standardly used in Gibbs sampling of HCRP.

\subsection{Overall sampling} \label{ss:generalsampler}


The Metropolis-Hastings algorithm 
 provides a way of constructing an MCMC sampler using unnormalized probability value $\tilde p(z)$.
After sampling $z^*$ from proposal distribution $q(z^*|z)$,
the algorithm computes acceptance probability $R$:
\begin{align}
R &{}= \min\, \biggl( 1, \, \dfrac{\tilde p(z^*)}{\tilde p(z^{old})} \dfrac{q(z^{old}|z^*)}{q(z^*|z^{old})} \biggr) \quad .
\end{align}
Then the result $z^{new} = z^*$ with probability $R$, and $z^{new} = z^{old}$ otherwise. 
Repeating this process constitutes an MCMC sampler from required distrubution $p(z) \propto \tilde p(z)$,



Within the context of HCRP, sample space $\mathbf{z}$ consists of draws $\mathbf{x}$ and seating arrangement $\mathbf{s}$.
From Eq.~\eqref{eq:drawcond}, we can use the non-restricted probability of draws $p( \mathbf{x}, \mathbf{s} )$
as unnormalized probability value \markerg{$\tilde p(\mathbf{z})$}, but it is not easy to provide
a proposal for joint distribution $q(\mathbf{x}^*, \markerg{\mathbf{s}^*})$.

Our idea is to factorize the proposal distribution as:
 \begin{align}
 	q(\mathbf{x}^*, &\mathbf{s}^*  | \mathbf{s}_{0}) 
 	   = q_{\mathbf{x}}(\mathbf{x}^* | \mathbf{s}_{0}) 
 	   \cdot q_{\mathbf{s}}(\mathbf{s}^* | \mathbf{x}^{*}, \mathbf{s}_{0}) \  .
 \end{align}
First factor $q_{\mathbf{x}}$ is the proposal distribution of the draws. 
Second factor $q_{\mathbf{s}}$ is the proposal distribution of the seating arrangements given the proposal draws.
We use the result of the $\mathsf{addCustomer}$ procedure, which stochastically updates the seating arrangements,
as the proposal distribution of the seating arrangements.\jfootnote{addCustomer 'ªà–¾'È'­'Å'Ä'­'é'Ì'¾'ªA'Ç'±'©'çà–¾'µ'½'ç'¢'¢'à'Ì'©EEE}

\subsection{Computing Factors}
The following describes each factor in $R$ and its computation.

\paragraph{True Probability }
$p( \mathbf{x}, \mathbf{s} )$ 
in Eq.~\eqref{eq:drawcond} is the joint probability of all draws $x_{ji}$:
\jfootnote{'È'ñ'['©Aseating arrangement 'Æ'±'ê'Ü'Å'·'×'Ä'Ì draw 'Ì joint probability 'Æ'¢'¢'½'¢'Ì'¾'ª'¤'Ü'­Œ¾'¦'È'¢}
\begin{align}
p(\mathbf{x},\mathbf{s}) ={}& \prod_j p(s^{(j)}) \cdot p(s^{(0)}) \\[-3mm]
\noalign{where}
p( s^{(j)} ) {}={}& 
\frac{\Gamma(\alpha_0)}{ \Gamma(\alpha_0 + n_{j \cdot \cdot}) } \cdot \alpha_0^{m_{j \cdot}} \cdot \prod_t \Gamma(n_{jt\cdot}) \\
p( s^{(0)} ) {}={}& \frac{\Gamma(\gamma)}{ \Gamma(\gamma + m_{\cdot \cdot}) } \cdot \gamma^{K} \cdot \prod_t \Gamma(m_{\cdot k})   \quad ,
\end{align}
\markerc{and $\Gamma$ is the Gamma function.}
This is the product of the probabilities of seating arrangements \markerc{(Eqs.~\ref{eq:hcrp1} to~\ref{eq:hcrp2})} for each customer.

In practice, we only need to calculate probabilities that account for the change in seating from $\mathbf{s}_{0}$,
because the probability for unchanged customers is cancelled out through reducing the fraction in $R$.
Let $\mathbf{s}_{0}$ be the seating arrangement for the unchanged customers, then
\begin{align}
\frac{p(\mathbf{s}^*)}{p(\mathbf{s}^{old})} &{}= \frac{p(\mathbf{s}^*|\mathbf{s}_{0})}{p(\mathbf{s}^{old}|\mathbf{s}_{0})}
\quad .
\end{align}
In fact, $p(\mathbf{x}^*,\mathbf{s}^{*}|\mathbf{s}_{0})$ is easily calculated along with \textsf{addCustomer} operations:
\begin{align}
p(&\mathbf{x}^*, \mathbf{s}^*|\mathbf{s}_{0}) = {} \nonumber \\ &
                  p(x^*_1,\mathbf{s}^*_{1}| \mathbf{s}_{0} )\, p(x^*_2,\mathbf{s}^*_{2}| \mathbf{s}^*_{1} )
               \cdots\, p(x^*_{\markerg{L}}, \mathbf{s}^{*}| \mathbf{s}^*_{\markerg{L}-1} ) \,.
\label{eq:true2}
\end{align}
Here, \markerg{$p(x_\ell, \mathbf{s}_{\ell}| \mathbf{s}_{\ell-1} )$} is probability 
\markerg{${p( x_{j_\ell i_\ell}, t_{j_\ell i_\ell} | j_\ell, \mathbf{s}_{\ell-1} )}$} of obtaining seating arrangement 
\markerg{${\mathbf{s}_{\ell}}$} as a result of drawing a sample from restaurant $j$:
\begin{align}
p( x_{ji} = k&, t_{ji} = t | j, \mathbf{s} ) = \dfrac{1}{n_{j \cdot \cdot} + \alpha_0  } \times {}  \nonumber \\[-1mm] &
     \begin{cases}
        ~ n_{jtk} 
        		& n_{jt\cdot} \ge 1 \\
        ~ \alpha_0 \cdot \dfrac{m_{\cdot k}}{m_{\cdot \cdot} + \gamma} 
        		& n_{jt\cdot} = 0, m_{\cdot k} \ge 1 \\
        ~ \alpha_0 \cdot \dfrac{\gamma}{m_{\cdot \cdot} + \gamma} 
        		& n_{jt\cdot} = 0, m_{\cdot k} = 0
	 \end{cases}
	 \label{eq:trueprob}
\end{align}
The same applies to the calculation of $p(\mathbf{s}^{old}|\mathbf{s}_{0})$, which can be done along with \textsf{removeCustomer} operations.

\paragraph{Proposal Distribution of Draws} \label{ss:proposaldraw}
 $q(\mathbf{x})$ can be anything as long as it is ergodic within restriction $C$.
To increase the acceptance probability, however, 
it is preferable for the proposal distribution to be close to the true distribution.
We suggest that a good starting point would be to use a joint distribution 
composed of the predictive distributions of each draw, as has been done in 
the approximated Gibbs sampler \cite{beal2002infinite}:
\begin{align}
q_{\mathbf{x}}(\mathbf{x}) = \mathbb{I}[\mathbf{x} \in C]\, \displaystyle{ \prod_{i=1}^{\markerg{L}} p( x_{i} | \mathbf{s}_{0} )} \ .
\label{eq:proposal}
\end{align}
We will again discuss the proposal distribution of draws for the HCRP-HMM case in Section~\ref{s:hmmsampler}.  

\paragraph{Proposal Distribution of Seating Arrangements}
 $q_s(\mathbf{s}^*|\mathbf{x}, \mathbf{s}_{0} )$, 
is the product of the probabilities for each operation of adding a customer:
\begin{align}
q_{\mathbf{s}}(\mathbf{s}^*|\mathbf{x}^*, \mathbf{s}_{0}) = {}& 
                 q_{\mathbf{s}}(\mathbf{s}_{1}^*| x_1^*, \mathbf{s}_{0} )\, q_{\mathbf{s}}(\mathbf{s}_{2}^*| x_{2}^*, \mathbf{s}_{1}^* ) \nonumber \\ 
               & \cdots\, q_{\mathbf{s}}(\mathbf{s}^{*}| x_\ell^*, \mathbf{s}_{\ell-1}^* ) \ .
\end{align}
Here, ${ q_{\mathbf{s}}(\mathbf{s}_{\ell}| x_{\ell}, \mathbf{s}_{\ell-1} )} = {p( t_{j_\ell i_\ell} | x_{j_\ell i_\ell}, j_\ell, \mathbf{s}_{\ell-1} )}$, 
\markerg{i.e.,} the probability 
of obtaining seating arrangement 
${\mathbf{s}_{\ell}}$ as a result of the ${\mathsf{addCustomer}(x_{j_\ell i_\ell}, j_\ell, {\mathbf{s}_{\ell-1}})}$ operation.
\begin{align}
p( t_{ji}&{} = t |  x_{ji} = k, j, \mathbf{s} ) = 
\\  &
     \begin{cases}
	\dfrac{ n_{jtk} }{n_{j \cdot k} + \alpha_0 \frac{m_{\cdot k}}{m_{\cdot \cdot} + \gamma} \,}
      				& n_{jt \cdot} \ge 1 \land k_{jt} = k \\[.5mm]
	\dfrac{ \alpha_0 \frac{m_{\cdot k}}{m_{\cdot \cdot} + \gamma } }{n_{j \cdot k} + \alpha_0 \frac{m_{\cdot k}}{m_{\cdot \cdot} + \gamma} \,}
      				& n_{jt \cdot} = 0 \land m_{\cdot k} > 0 \\[.5mm]
      1
      				& n_{jt \cdot} = 0 \land m_{\cdot k} = 0
     \end{cases} \quad .
 \label{eq:proposeseat}
\end{align}

\subsection{Simplification}

Paying attention to the fact that both Eqs.~\eqref{eq:trueprob} and~\eqref{eq:proposeseat} are calculated along 
a series of $\mathsf{addCustomer}$ calls, 
we introduce factors
\begin{align}
r^*_\ell &{} = 
\frac{ p(x_\ell^*,\mathbf{s}_{\ell}|\mathbf{s}_{\ell-1}) }{ q_{\mathbf{s}}(\mathbf{s}_{\ell}|x_\ell^*, \mathbf{s}_{\ell-1}) }
&
r^{old}_\ell &{} = 
\frac{ p(x_\ell^{old},\mathbf{s}_{\ell}|\mathbf{s}_{\ell-1}) }{ q_{\mathbf{s}}(\mathbf{s}_{\ell}|x_\ell^{old}, \mathbf{s}_{\ell-1}) }
\label{eq:refactor}
\end{align}
to simplify the calculation of $R$ as:
\begin{align}
R &{} = \min\, \biggl( 1,\, \dfrac{{p(\mathbf{s}^*)}}{{p(\mathbf{s}^{\mathit{old}})}} 
		\dfrac{q_{\mathbf{s}}( \mathbf{s}^{\mathit{old}} | \mathbf{x}^{\mathit{old}}, \mathbf{s}_{0})}{q_{\mathbf{s}}(\mathbf{s}^{*} | \mathbf{x}^{*}, \mathbf{s}_{0})}
		\dfrac{q_{\mathbf{x}}(\mathbf{x}^{\mathit{old}})}{q_{\mathbf{x}}(\mathbf{x}^{*})}  \, \biggr)  \nonumber \\
   &{} = \min\, \biggl( 1,\, 
   		\dfrac{ r( \mathbf{x}^*,\mathbf{s}^*|\mathbf{s}_{0} ) }{ r( \mathbf{x}^{old},\mathbf{s}^{old}|\mathbf{s}_{0} )  }
		\dfrac{{q(\mathbf{x}^{\mathit{old}})}}{q(\mathbf{x}^{*})}  \, \biggr)  \quad ,
\end{align}
where
\begin{align}
r( \mathbf{x}^*,\mathbf{s}^*|\mathbf{s}_{0} ) 
&{} = \frac{ p(\mathbf{x}^*,\mathbf{s}^*|\mathbf{s}_{0}) }{ q_{\mathbf{s}}(\mathbf{s}^*|\mathbf{x}^*, \mathbf{s}_{0}) } \nonumber \\
&{} = 
\frac{ p(x_1^*,\mathbf{s}_{1}|\mathbf{s}_{0}) }{ q_{\mathbf{s}}(\mathbf{s}_{1}|x_1^*, \mathbf{s}_{0}) } 
\cdots
\frac{ p(x_L^*,\mathbf{s}_{L}|\mathbf{s}_{L-1}) }{ q_{\mathbf{s}}(\mathbf{s}_{L}|x_L^*, \mathbf{s}_{L-1}) }  \nonumber \\
&{} = \markerg{r_1^* \cdot r_2^*  \cdots r_L^*} \quad .
\end{align}

Surprisingly, assigning Eqs.~\eqref{eq:trueprob} and~\eqref{eq:proposeseat} into Eq.~\eqref{eq:refactor}
reveals that \markerg{$r_\ell^*$} is equal to $p(\markerg{x_{j_\ell i_\ell}} = x_\ell^*|\mathbf{s}^*_{\ell-1} )$,
i.e., the probability of new customer \markerg{$x_{j_\ell i_\ell}$} at restaurant $j_\ell$ eating dish $x_\ell^*$:
\begin{align}
p( x_{ji} = k | \mathbf{s}) &{}= \frac{ \, n_{j \cdot k} + \alpha_0 \frac{m_{\cdot k}}{m_{\cdot k} + \gamma} \, } { n_{j \cdot \cdot} + \alpha_0 } 
\label{eq:predictive1} \\
p( x_{ji} = k^{new} | \mathbf{s}) &{}= 
	\frac{ \, \alpha_0 \frac{\gamma}{m_{\cdot k} + \gamma} \,} { n_{j \cdot \cdot} + \alpha_0 }  \quad .
\label{eq:predictive2}
\end{align}
In other words, calculation of the accept ratio does not use $t_{ji}$ (the table index of each customer), 
despite the fact that the values of $t_{ji}$ are being proposed; $t_{ji}$ will indirectly affect the accept ratio by 
changing subsequent draw probabilities $p(x_{\ell+1}^*|\mathbf{s}^*_{\ell}), p(x_{\ell+2}^*|\mathbf{s}^*_{\ell+1}), \ldots$  
through modifying $n_{jtk}$ and $m_{jk}$, i.e., the number of customers and tables.  

It is now clear that, as done in 
some previous work \cite{teh2006bayesian}, we can save storage space by using 
an alternative representation for seating arrangements $\mathbf{s}$, in which the table indices 
of each customer $t_{ji}$ are forgotten but only the numbers of customers $n_{jt\cdot}$, $k_{jt}$ and $m_{jk}$ are retained.
The only remaining reference to $t_{ji}$ in the $\mathsf{removeCustomer}$ procedure can be safely replaced by sampling.

However, it should be noted that we have to revert to original
seating assignment $\mathbf{s}^{old}$ whenever the proposal is rejected.  
Putting the old draws $\mathbf{x}^{old}$ back into $\mathbf{s}_{0}$ by using the $\mathsf{addCustomer}$ procedure again
will lead sampling to an incorrect distribution of seating assignments, and consequently, 
an incorrect distribution of draws.

\begin{algorithm*}
\caption{\textsf{MH-RCDSampler}($\mathbf{j}$, $\mathbf{x}^{old}, \mathbf{s}^{old}$): Metropolis-Hastings sampler for restricted collapsed draw}
\label{alg:mhcsd}
\begin{algorithmic}[1]
\STATE ${\mathbf{s}_{L}^{\markerg{old}}} = \mathbf{s}^{old}$
\FOR{ $\ell=L$ \DOWNTO $1$ }
\STATE ${\mathbf{s}_{\ell-1}^{\markerg{old}}} = \mathsf{removeCustomer}( x^{old}_\ell, j_\ell, \mathbf{s}_{\ell}^{\markerg{old}}  )$
\COMMENT{Remove customers for $x^{old}_1, \ldots, x^{old}_{m}$ sequentially from $\mathbf{s}^{old}$}
\label{algline:remove}
\STATE $r^{old}_\ell = p(\x^{old}_\ell, \mathbf{s}_{\ell-1}^{\markerg{old}})$
\COMMENT{Calculate factors for accept ratio}
\ENDFOR
\STATE ${\mathbf{s}_{0}^*} = {\mathbf{s}_{0}} = \mathbf{s}_{0}^{\markerg{old}}$
\STATE $\mathbf{x}^* \sim q_{\mathbf{x}}(\mathbf{x}; \mathbf{s}_{0})$  \COMMENT{Draw $\mathbf{x}^*$ from proposal distribution $q(\mathbf{x})$ of draws.}
\label{algline:draw}
\FOR{ $\ell=1$ \TO $L$ }
\STATE $r^{*}_\ell = p(x^{*}_\ell, \mathbf{s}_{\ell-1}^*)$
\COMMENT{Calculate factors for accept ratio}
\STATE          $\mathbf{s}_{\ell}^* = \mathsf{addCustomer}(x^*_{\ell}, j_\ell, \mathbf{s}_{\ell-1}^*)$
                                    \COMMENT{Add customers for $x^*_{1}, \ldots, x^*_{m}$ sequentially to $\mathbf{s}^*_{0}$}
\label{algline:seat1}
\ENDFOR                             					
\STATE   $\mathbf{s}^* = \mathbf{s}_{L}^*$             \COMMENT{Obtain proposal seating $\mathbf{s}^*$}
\label{algline:seat2}
\STATE  $\displaystyle R = \min\, \biggl( 1,\, \dfrac{{\markerg{q_{\mathbf{x}}}(\mathbf{s}^{\markerg{old}})}}{{\markerg{q_{\mathbf{x}}}(\mathbf{s}^{\markerg{*}})}} 
			\prod_{\ell=1}^L \dfrac{r^*_\ell}{r^{old}_\ell}  \, \biggr) $
		                            \COMMENT{Calculate acceptance probability}
\label{algline:calcr}
\RETURN $\langle \mathbf{x}^{\mathit{new}}, \mathbf{s}^{\mathit{new}} \rangle =
\begin{cases}
~\langle \mathbf{x}^*, \mathbf{s}^* \rangle				& \mbox{with probability $R$} \\
~\langle {\mathbf{x}^{\mathit{old}}, \mathbf{s}^{\mathit{old}}} \rangle	& \mbox{otherwise.}
\end{cases}
$
		                            \COMMENT{Accept/reject proposed sample}
\label{algline:accept}
\end{algorithmic}
\end{algorithm*}

Algorithm~\ref{alg:mhcsd} is the one we propose othat obtains new samples ${\mathbf{x}^{\mathit{new}}}$ 
drawn simultaneously from restaurants indexed by $\mathbf{j}$ and associated seating arrangement ${\mathbf{s}^{\mathit{new}}}$,
given previous samples ${\mathbf{x}^{\mathit{old}}}$ and ${\mathbf{s}^{\mathit{old}}}$.

The first half of this sampler is similar to a sampler for a single draw;
it consists of removing old customers (line~\ref{algline:remove}), 
choosing a new sample (line~\ref{algline:draw}), and
adding the customers again (line~\ref{algline:seat1}).  
The main difference is that there are $L$ times of iteration for each call $\mathsf{removeCustomer}$/$\mathsf{addCustomer}$,
and the calculation of $r$, which is later used for acceptance probability $R$.

\section{Gibbs sampler for HCRP-HMM} \label{s:hmmsampler}

This section describes a series of samplers for HCRP-HMM.
First, we present the step-wise Gibbs sampler as the simplest example.
After that, we describe a blocked Gibbs sampler using a forward-backward algorithm.
We also explain the HCRP version of the beam sampler \cite{vangael2008beam} as well as
the split-merge sampler \cite{jain2000split} for iHMM.

\subsection{Step-wise Gibbs sampler}
A step-wise Gibbs sampler for HCRP-HMM is easily constructed using an RCD sampler (Algorithm~\ref{alg:gibbs} \markerg{in the Appendix} describes one Gibbs sweep).
We slightly modified the proposal distribution $q(x_t)$ from that suggested in Section~\ref{ss:proposaldraw},
in order to ensure that $x_{t+1}$ is proposed with non-zero probability even when no table in $\mathbf{s}_{0}$ serves dish $x_{t+1}$:
\begin{align}
  q_{\markerg{x}}(x_t) \propto  {}&
  \!\!\left( \!p( x_t | s_{0}^{(x_{t-1})} ) + \!\left(\!\frac{\alpha_0 \gamma}{(\alpha_0 + n_{x_{t-1} \cdot \cdot}) (\gamma + m_{\cdot \cdot})} \!\right) \! \delta_{x_{t+1}} \! \right)
  \nonumber\\ & \cdot   p( x_{t+1} | s_{0}^{(x_{t})} )  \cdot  p( y_t | F_{0}^{(x_t)} )
  \quad .
\end{align}

\subsection{Blocked Gibbs sampler} \label{ss:blockgibbs}

We can construct an alternate sampling scheme under the framework of RCD sampler 
that resamples a block of hidden states simultaneously, based on the forward-backward sampler \cite{scott2002bayesian}.
The idea is that we run the forward-backward sampler with a predictive transition distribution from HCRP-HMM, and
use the result as a proposal of restricted collapsed draw.

For iHMM, the forward-backward sampling algorithm \cite{scott2002bayesian} cannot be directly used, because
the forward probability values for an infinite number of states have to be stored for each time step $t$  \cite{vangael2008beam}.
This is not the case for HCRP-HMM, 
because predictive transition probability $\mathbf{\hat \pi}$ from given seating assignment $\mathbf{s}_{0}$,
which is given as Eqs.~\eqref{eq:predictive1} and~\eqref{eq:predictive2},
only contains transition probability for finite number $K$ of states plus one for $k^{new}$.
Thus we only need to store $K+1$ forward probability for each time step $t$.

Result $\mathbf{\bar x}$ of the forward-backward sampler, however, cannot be used directly as the proposal;
\markerg{the $i$-th state of the proposal $x^*_i$ is equal to $\bar x_i$ when $\bar x_i \ne k^{new}$, but }
we need to assign new state indices to $x^*_i$ whenever $\bar x_i = k^{new}$.
In particular, when $k^{new}$ has appeared $W \ge 2$ times, 
all appearances of $k^{new}$ may refer either to the same new state, or to $W$ different states, or 
to anything in between the two, in which some appearances of $k^{new}$ share a new state.
\jfootnote{'·'ׂĂ̏ó'ÔŒn—ñ'ª non-zero probability 'Å'Å'Ä'­'é'悤'É'·'é'ɂ́A
forward-backward 'ŃTƒ"ƒvƒ‹'µ'½ó'ÔŒn—ñ'É•¡"'Ì $k^{new}$ 'ªŠÜ'Ü'ê'éê‡'ɁA
'Ç'Ì $k^{new}$ 'Æ'Ç'Ì $k^{new}$ '𓯂¶ index 'É'µ'āA'Ç'Ì $k^{new}$'ðˆá'¤ index 'É'·'é'©A'Ì'·'ׂẴpƒ^[ƒ"'ª
propose '³'ê'é'悤'É'·'é•K—v'ª' 'éB'Æ'¢'¤'Ì'ð‰pŒê'ŏ''«'½'¢'ª'¤'Ü'­''¯'È'¢``'½'·'¯'ā````}

To achieve this purpose, we prepare special CRP $Q^*$ that accounts for the previously unseen states, 
marked by $k^{new}$ in the result of the forward-backward sampler.
Specifically, each table in $Q^*$ has a dish with an unused state index, and 
each appearance of $k^{new}$ is replaced with a draw from $Q^*$.
This construction ensures that every state sequence is proposed with a non-zero probability, and 
allows the proposal probability to be easily calculated.
The concentration parameter of $Q^*$ is set as equal to $\gamma$.
To handle the case where some of the new states are equal to $x_{t_{b+1}}$, i.e., index of the state that succeeds to the resampling block, 
we add to $Q^*$ an extra customer that correponds to $x_{t_{b+1}}$
when $x_{t_{b+1}}$ does not appear in $\mathbf{s}_{0}$,

Resulting proposal probability is:
\begin{align}
\markerg{q_{\mathbf{x}}}&\markerg{(\mathbf{x}^*)} =  {} \nonumber \\ &
\left( \prod_{\ell=0}^L \hat \pi_{\bar \x_{\ell+1} \bar \x_{\ell}} \cdot 
				  \prod_{\ell=1}^L F_{\bar \x_{\ell}}(y_\ell) \right)  \cdot 
				  \prod_{\ell:\bar \x_\ell = k^{new}} p( \x_\ell^* | Q^* ) \quad ,
\end{align}
where the first factor accounts for the forward probability of the sequence, 
and the second factor accounts for probability of the new state assignment.

Note also that, to make a sensible proposal distribution, we cannot resample the whole state sequence simultaneously.  We need to 
divide the state sequence into several blocks, and resample each block given the other blocks.  The size of a block affects
efficiency, because blocks that are too large have lower accept probability, while with blocks that are too small, the algorithm has little advantage over
step-wise Gibbs sampling.  

Algorithm~\ref{alg:blockgibbs} in the Appendix describes one sweep of a blocked Gibbs sampler for an HCRP-HMM\@.  

\subsection{Beam sampling}

Beam sampling for HDP-HMM \cite{vangael2008beam} is a sampling algorithm 
that uses slice sampling \cite{neal2003slice} for transition probability to extract
a finite subset from the state space.
Although the possible states are already finite in HCRP-HMM, the same technique may benefit sampling of HCRP-HMM 
by improving efficiency from the reduced number of states considered during one sampling step.

We just need replace the call to $\mathsf{ForwardBackwadSampling}$ in Algorithm~\ref{alg:blockgibbs}
with the call to $\mathsf{BeamSampling}$ to use beam sampling with HCRP-HMM.
A brief overview of the beam sampling is:
\begin{enumerate}
\item Sample auxiliary variables $\mathbf{u}=(u_0, \ldots, u_L)$  as $u_\ell \sim \mathrm{Uniform}(0, \pi_{\x_{\ell}\x_{\ell-1}})$,
\item  For $\ell=1, \ldots, L$, calculate forward probability $q(x'_\ell = k')$ using a slice of transition probability 
      \mbox{$q(x'_{\ell} = k') = F_{k'}(y_{\ell}) \sum_k \mathbb{I}( \pi_{k'k} > u_{\ell-1} ) q(x'_{\ell-1} \!= k)$,}
\item  For $\ell=L, \ldots, 1$, sample the states $x'_\ell$ backwardly, i.e. $p(x'_\ell = k) \propto \mathbb{I}( \pi_{x'_{\ell+1}k} > u_\ell ) $. 
\end{enumerate}
For details, refer to the original paper \cite{vangael2008beam}.  

Some remarks may be needed for the calculation of $q^*_{\mathbf{\markerg{x}}}$, i.e., the proposal probability for the state sequence.
Although beam sampling has a different proposal distribution from forward-backward sampling, 
we can use the same calculation of proposal probability used in acceptance probability as
that of forward-backward sampling. This is because beam sampling
satisfies the detailed balance equation, which ensures that 
the ratio of proposal probability with beam sampling $\frac{q^*_{slice}}{q^{old}_{slice}}$ is
always equal to the ratio of the probability obtained by forward-backward sampling $\frac{q^*}{q^{old}}$.

\subsection{Split-Merge Sampling}

We can integrate the split-merge sampling algorithm, which is another sampling approach to Dirichlet process mixture models \cite{jain2000split}, 
into HCRP-HMM using the RCD sampler.  
A split-merge sampler makes a proposal move that tries to merge two mixture components into one, or to split a mixture component into two; 
the sampler then uses a Metropolis-Hastings step to stochastically accept the proposal.  
Based on the RCD framework, we can extend the split-merge sampler into HCRP,
which can be applied to HCRP-HMM\@.  Within the context of HMM, the sampler corresponds to merge two state indices into one, or to split a state index into two.  

Our implementation is based on an improved version of hte split-merge sampler, called the sequentially-allocated merge-split sampler 
\cite{dahl2005sequentially-allocated}, which produces a split proposal while sequentially allocating components in random order.
To deal with temporal dependency in HMM, we identify fragments of state sequences to be resampled within the state sequence,
and perform blocked Gibbs sampling for each fragment in random order.

We added one important optimization to the split-merge sampling algorithm.
Since a merge move is proposed much more frequently than a split move, and the 
move has a relatively low accept probability,
it is beneficial if we have a way of determining whether a merge move is rejected or not earlier.
Let us point out that, when proposal probability for a merge move is calculated, 
the accept probability is monotonically decreasing.
Consequently we sample $R^{thr}$, the threshold of accept probability, at the beginning of the algorithm and
stop further calculation when $R$ becomes less than $R^{thr}$.
Algorithm~\ref{alg:splitmerge}
 in the Appendix is the split-merge sampling algorithm for HCRP-HMM\@.

Split-merge sampling allows faster mixing when it is interleaved with other sampling strategies.
We examine split-merge sampling with each of the samplers we have presented in this paper.

\section{Experiments and Discussion}

This section presents two series of experiments, the first with small artificial sequences and the second with a sequence of natural language words.

\subsection{Settings}

We put gamma prior $\mathrm{Gamma}(1,1)$ on $\alpha_0$ and $\gamma$, and sampled 
between every sweep using an auxiliary variable method  \cite{teh2006hierarchical} in all the experiments.
We introduced HCRP as a prior of emission distributions as well, and its hyperparameters were also sampled in the same way.

The initial state sequence given to the sampler is the result of a particle filter with 100 particles.

We measured autocorrelation time (ACT) to evaluate mixing.
Given a sequence of values $\mathbf{x} = x_1, x_2, \ldots, x_T$, its mean $\mu$ and variance $\sigma^2$, 
$ACT(\mathbf{x})$ are defined as follows:
\begin{align}
ACF_t(\mathbf{x}) &{} = \frac{1}{(T-t) \sigma^2} \sum_{i=1}^{T-t} (x_i - \mu) (x_{i+t} - \mu) 
\\ 
ACT(\mathbf{x}) &{} = \frac{1}{2} + \sum_{t=1}^{\infty} ACF_t(\mathbf{x}) \qquad .
\end{align}
Since with larger $t$, $ACF_\ttt(\mathbf{x})$ is expected to converge to zero, we used $ACF_\ttt(\mathbf{x})$ for $t \le 1000$.

For artificial sequence, we evaluated mutual information between the $h_t$, hidden state used in sequence generation
and $\x_t$, inferred states as follows:
\begin{align}
MI &{}= \sum_h \sum_x p(x, h) \log \frac{p(x, h)}{p(x) p(h)} \quad .
\end{align}
For natural language text, the inferred model is evaluated by multiple runs of a particle filter 
on a given test sequence of length $T_{test}$. We specifically construct a particle filter with $Z=100$ particles 
for each sampled model state $\mathbf{s}_z$, and 
evaluate likelihood $l(y_\ttt|\mathbf{s}_z)$ for each emission. 
Finally, we calculate the perplexity (the reciprocal geometric mean of the emission probabilities) of the test sequence:
\begin{align}
\nonumber \\[-7.5mm]
PPL &{}= \exp \left( - \frac{1}{T_{test}} \sum \log \hat l(y_\ttt) \right) \quad \\
\noalign{where}
\hat l(y_\ttt) &{} = \frac{1}{Z} \sum_{z=1}^{\markerg{Z}} l( y_\ttt | \mathbf{s}_{z})   \quad 	.
\end{align}

The samplers we chose for comparison are 
the step-wise Gibbs sampler with direct assignment representation \cite{teh2006hierarchical}, which uses
stick-breaking for the root DP and CRP for the other DPs,
the step-wise Gibbs sampler with stick-breaking construction, and
the beam sampler with stick-breaking construction \cite{vangael2008beam}.
\marker{For} fair comparison between different algorithms, we collected samples to evaluate the autocorrelation time and perplexity
on a CPU-time basis (excluding the time used in evaluation). 
All the algorithms were implemented with C++ and tested on machines with an Intel Xeon E5450 at 3~GHz.

\subsection{Artificial data}

\begin{figure}
\centering{\includegraphics[width=8cm]{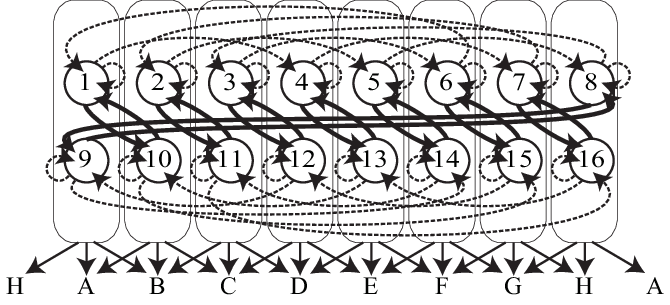}}
\caption{Automaton that generates Sequence 2\@. Circles denote hidden states, and the same alphabet emissions are observed from states within an oval group. 
A dashed arrow denotes transition with probability 0.8,
a bold arrow denotes transition with probability 0.84, and
a solid arrow denotes  emission with probability 1/3.
  } \label{fig:automaton}
\end{figure}

The first series of experiments are performed with two small artificial sequences.
Sequence 1 consists of repeating sequence of symbols A-B-C-D-B-C-D-E-... for length $T=500$, 
and we run the sampler 30~s for burn-in, and after that, a model state is sampled every 2~s until a total of 300~s is reached.
Sequence 2 is generated from the simple finite state automaton in Figure~\ref{fig:automaton} for length $T=2500$,
and we use 60~s for burn-in and total 600~s.
We evaluated the mutual information between the inferred hidden states and the true hidden states.

\begin{figure*}
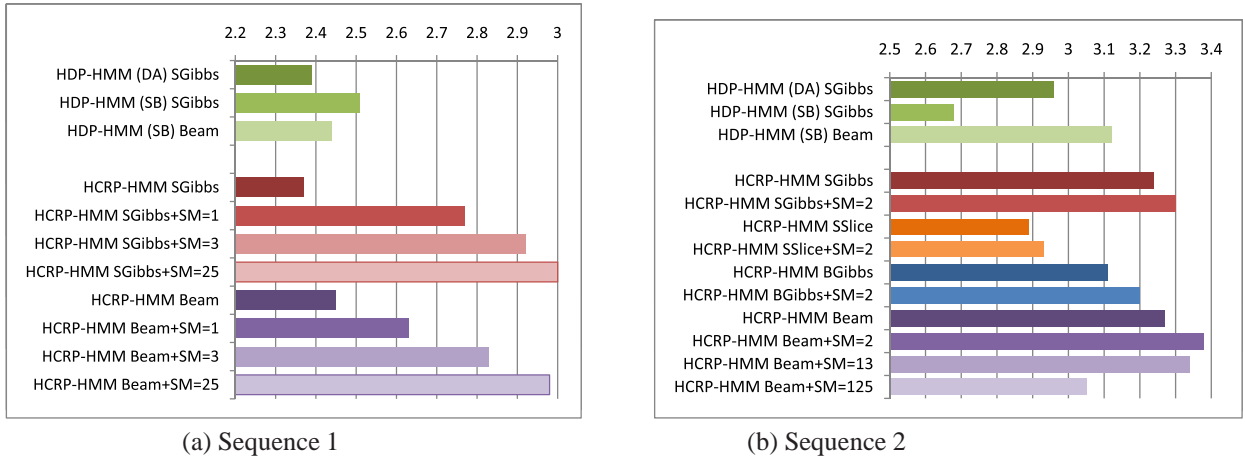

\begin{tabular}{cc}
\includegraphics[width=6cm]{graph-p2.epsi} \qquad\qquad & 
\qquad\qquad \includegraphics[width=6cm]{graph-p6.epsi} \\
(a) Sequence 1 & (b) Sequence 2
\end{tabular}
\caption{Average mutual information of sampled hidden states}\label{fig:graph1}
\end{figure*}

Figure~\ref{fig:graph1} shows the distribution of mutual information for 100 trials after 300 s.
We can see that some of the samplers based on the proposed method achieved a better mutual information compared to existing samplers.
The improvement depends on the type of sequence and the samplers.

\begin{figure*}
\begin{tabular}{ccc}
\includegraphics[trim=20 0 0 0]{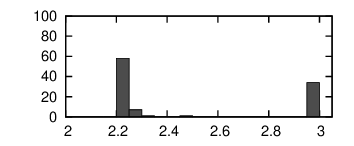} & 
\includegraphics[trim=20 0 0 0]{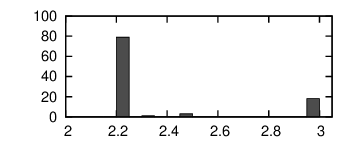} &
\includegraphics[trim=20 0 0 0]{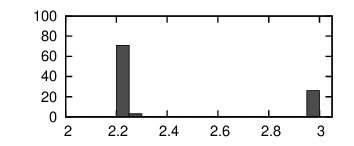} \\
(a) HDP-HMM (SB) SGibbs & (b) HDP-HMM (DA) SGibbs & (c) HDP-HMM (SB) Beam \\[1mm]
\includegraphics[trim=20 0 0 0]{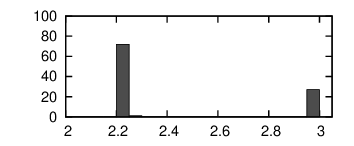} &
\includegraphics[trim=20 0 0 0]{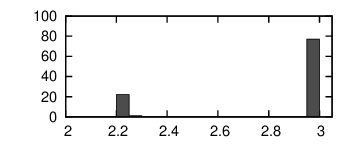} &
\includegraphics[trim=20 0 0 0]{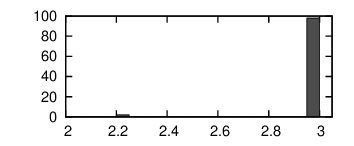}  \\
(d) HCRP-HMM Beam & (e) HCRP-HMM Beam & (f) HCRP-HMM Beam		\\
                  & Split-Merge 3/sweep & Split-Merge 25/sweep
\end{tabular}
\caption{Distribution of mutual information for Sequence 1.  X-axis shows mutual information and
Y-axis shows frequency.  Block size $\approx 6$ for HCRP-HMM Beam sampling. }\label{fig:graph}
\end{figure*}

For Sequence 1, we can see that split-merge sampling yields better results compared to other samplers. 
Although HMM with eight hidden states can completely predict the sequence, 
the samplers tend to be trapped in a local optimum with five states in the initial phase,
because our selected prior of $\gamma$ poses a larger probability on a smaller number of hidden states,
Detailed investigations (Figure~\ref{fig:graph}) confirmed this analysis.

For Sequence 2, on the other hand, blocked samplers worked very efficiently.  
Step-wise samplers generally worked poorly on the sequence, because the strong dependency on temporally adjacent states impedes mixing.
  Still, step-wise Gibbs sampler for HCRP-HMM outperformed the beam sampler with the stick-breaking process. 
The blocked Gibbs sampler had inferior performance due to its heavy computation for a large number of states, 
but the beam sampler for HCRP-HMM was efficient and performed well.  Combination with
a small number of split-merge samplers increases the performance (more split-merge sampling 
leads to lower performance by occupying computational resource for the beam sampler).
\markerf{From averages statistics of samplers (Table~\ref{tbl:seq2}), we can see that 
 (1) the increase of mutual information cannot be described only by the increase of the number of states;
 (2) The accept ratio for the Gibbs trial has a very high accept rate;
 (3) Split-merge samplers have a very low accept rate, but still make improvement for mutual information. }

\begin{table*}
\caption{Experimental results for Sequence 2} \label{tbl:seq2}
\markerf{
\centering{
\begin{tabular}{l|rrrrrr}
name & MI & ACT \#states & \#states & secs/sweep & Gibbs accept \markerg{rate} & SM accept \markerg{rate}\\
\hline
HDP-HMM (DA) SGibbs     &    2.92 &   0.527 &  14.910 &   0.044 & --- & --- \\ 
HDP-HMM (SB) Beam       &    3.04 &   0.640 &  14.720 &   0.032 & --- & --- \\ 
HCRP-HMM SGibbs         &    3.18 &   0.719 &  16.210 &   0.026 & 0.999666 & --- \\ 
HCRP-HMM SGibbs +SM=2   &    3.28 &   0.615 &  17.190 &   0.030 & 0.999632 & 0.000631 \\ 
HCRP-HMM SGibbs +SM=13  &    3.19 &   0.493 &  16.830 &   0.044 & 0.999619 & 0.000650 \\ 
HCRP-HMM SSlice         &    2.82 &   0.820 &  15.950 &   0.009 & 0.999847 & --- \\ 
HCRP-HMM SSlice +SM=2   &    2.86 &   0.705 &  17.330 &   0.013 & 0.999822 & 0.000827 \\ 
HCRP-HMM SSlice +SM=13  &    2.59 &   0.604 &  16.400 &   0.030 & 0.999830 & 0.000910 \\ 
HCRP-HMM BGibbs         &    3.01 &   0.317 &  14.900 &   0.206 & 0.995525 & --- \\ 
HCRP-HMM BGibbs +SM=2   &    3.12 &   0.513 &  16.270 &   0.237 & 0.995135 & 0.000985 \\ 
HCRP-HMM BGibbs +SM=13  &    3.18 &   0.637 &  16.530 &   0.265 & 0.994715 & 0.000715 \\ 
HCRP-HMM Beam           &    3.21 &   0.866 &  15.180 &   0.016 & 0.997233 & --- \\ 
HCRP-HMM Beam +SM=2     & \textbf{3.37} & 0.898 &  16.910 &   0.019 & 0.996369 & 0.000497 \\ 
HCRP-HMM Beam +SM=13    &    3.28 &   0.875 &  17.070 &   0.034 & 0.996316 & 0.000532 \\ 
\end{tabular}
}
}
\small
\centering{
\begin{tabular}{ll}
DA: Direct Assignment &
SB: Stick-Breaking construction \\
\markerf{MI: Mutual Information} &
\markerf{ACT: Auto-correlation time, samples collected for every 0.1 s } \\
\markerf{\#states: number of states} & SGibbs: step-wise Gibbs \\
\multicolumn{2}{l}{SSlice: step-wise Gibbs with slice sampling (beam sampling with block size=1)} \\
\multicolumn{2}{l}{BGibbs: blocked Gibbs (block size $\approx$ 8)} \\
\multicolumn{2}{l}{Beam: beam sampling (block size $\approx$ 8 for HCRP-HMM, $T$ for stick-breaking)} \\
\multicolumn{2}{l}{SM: Split-Merge sampler (+SM=$n$ denotes SM trials per Gibbs sweep)}\\
\end{tabular}
}
\end{table*}

\subsection{Natural language text}

We also tested the samplers using a sequence of natural language words from \textit{Alice's Adventure in Wonderland}.
We converted the text to lower case, removed punctuation, and placed a special word EOS after every sentence to obtain 
a corpus with $28,120$ words; we kept the last 1,000 words for test corpus and learned on a sequence
with length $T=27120$.
We introduce a special word UNK (unknown) to 
replace every word that occurred only once, resulting in $|\Sigma| = 1,487$ unique words in the text.
We took 10,000~s for burn-in, and sampled a model state for every 120~s, until the total of 172,800~s.
Table~\ref{tbl:alice} summarize the averaged statistics for 18 trials. 

\begin{table*}
\caption{Experiments on Natural language text}\label{tbl:alice}
\centering{
\begin{tabular}{l|rrrrrr}
Sampler  &  Perplexity  &   \# states  & sec/sweep & Gibbs accept rate & SM accept rate \\
\hline
HDP-HMM (DA) SGibbs     &  134.22 & 313.056 &  10.017 & --- & --- \\ 
HDP-HMM (SB) SGibbs     &  151.10 & 242.389 &  38.045 & --- & --- \\ 
HDP-HMM (SB) Beam       &  178.59 &  68.444 &  16.126 & --- & --- \\ 
HCRP-HMM SGibbs         &  133.31 & 379.833 &   7.027 & 0.999861 & --- \\ 
HCRP-HMM SGibbs+SM=130  &  131.66 & 386.833 &   7.664 & 0.999857 & 0.000052 \\ 
HCRP-HMM SGibbs+SM=5400 &  135.94 & 336.278 &  31.751 & 0.999880 & 0.000050 \\ 
HCRP-HMM SSlice         &  131.17 & 422.000 &   0.469 & 0.999986 & --- \\ 
HCRP-HMM SSlice+SM=130  &  131.67 & 409.833 &   0.976 & 0.999993 & 0.000052 \\ 
HCRP-HMM SSlice+SM=5400 &  152.76 & 254.722 &  36.617 & 0.999993 & 0.000056 \\ 
HCRP-HMM BGibbs         &  199.14 & 1840.833 & 29380.261 & 0.992748 & --- \\ 
HCRP-HMM Beam           &  141.77 & 603.333 &  80.681 & 0.995627 & --- \\ 
HCRP-HMM Beam+SM=130    &  142.69 & 567.278 &  72.217 & 0.995612 & 0.000124 \\ 
HCRP-HMM Beam+SM=5400   &  141.07 & 495.667 &  84.925 & 0.995554 & 0.000101 
\end{tabular}
}
\small
\centering{
\begin{tabular}{ll}
For HCRP-HMM, the block size $\approx 10$. 
\end{tabular}
}
\end{table*}

We found that step-wise sampling outperformed blocked sampling (including beam sampling).
The reason for this may be the nature of the sequence, which has a lower temporal dependency.
Blocked Gibbs sampling, in particular, consumes too much time for one sweep to be of any practical use.
We also found that split-merge sampling had a very low accept rate and thus made little contribution to the result.

Yet, we can see the advantage of using HCRP representation over stick-breaking representation.  
The direct assignment (DA) algorithm showed a competitively good perplexity, reflecting the fact that DA uses stick-breaking
for only the root DP and uses the CRP representation for the other DP\@. Though step-wise Gibbs
sampling and its slice sampling version seems outperforming DA slightly, we need to collect more data to 
show that the difference is significant.  At least, however, we can say that now many sampling algorithms
are available for inference, and we can choose a suitable one depending on the nature of the sequence.

\section{Conclusion and Future Work}

We have proposed a method of sampling directly from constrained distributions of simultaneous draws 
from a hierarchical Chinese restaurant process (HCRP).
We pointed out that, to obtain a correct sample distribution, the seating arrangements (partitioning) must be correctly sampled
for restricted collapsed draw, and we thus proposed applying the Metropolis-Hastings algorithm to the seating arrangements.
Our algorithm, called the Restricted Collapsed Draw (RCD) sampler, 
uses a na\"\i{}ve sampler to provide a proposal distribution for seating arrangements.
Based on the sampler, we developed various sampling algorithms for HDP-HMM based on HCRP representation,
including blocked Gibbs sampling, beam sampling, and split-merge sampling.

The applications of the RCD sampler, which is at the heart of our algorithms, 
are not limited to HCRP-HMM\@.  The experimental results revealed that 
some of the proposed algorithms outperform existing sampling methods,
indicating that the benefits of using a collapsed representation exceed the cost of rejecting proposals.

The main contribution of this study is that it opens a way of developing more complex Bayesian models based on CRPs.
Since the RCD sampler is simple, flexible, and independent of the particular structure of a hierarchy, 
it can be applied to any combination or hierarchical structure of CRPs. 
Our future work includes using this algorithm to construct new Bayesian models based on hierarchical CRPs, 
which are hard to implement using a non-collapsed representation.
Planned work includes extending HDP-IOHMM \cite{doshivelez2009infinite} with a three-level hierarchical DP 
(e.g., the second level could correspond to actions, and the third level, to input symbols).

%

\bibliographystyle{plainnat}
\bibliography{makino}

\clearpage

\appendix

\section{Miscellaneous Algorithms}

\begin{algorithm}[h]
\caption{$\mathsf{getProb}(j, k, \mathbf{s})$: Calculate $p(x_{ji} = k|\mathbf{s})$ }
\label{alg:getProb}
\begin{algorithmic}
\IF{ $m_{\cdot k} = 0$ }
	\RETURN $\frac{ \, \alpha_0 \frac{\gamma}{m_{\cdot k} + \gamma} \,} { n_{j \cdot \cdot} + \alpha_0 }  $
\ELSE
	\RETURN $\frac{ \, n_{j \cdot k} + \alpha_0 \frac{m_{\cdot k}}{m_{\cdot k} + \gamma} \, } { n_{j \cdot \cdot} + \alpha_0 } $
\ENDIF
\end{algorithmic}
\end{algorithm}

\begin{algorithm}[h]
\caption{\textsf{addCustomer}($j, k, \mathbf{s}^{old}$): Adds new customer eating dish $k$ to restaurant $j$. }
\label{alg:addCustomer}
\begin{algorithmic}
\STATE $\mathbf{s} := \mathbf{s}^{old}$
\STATE
	With probabilities proportional to:
		$n_{jtk}\ (t = 1, \ldots, m_{j \cdot})$: Increment $n_{jtk}$ (the customer sits at $t$-th table)
		$\alpha_0 \frac{m_{\cdot k} }{m_{\cdot \cdot}+\gamma}$: 
			sit customer at a new table $t^{new}$ serving dish $k$ in restaurant $j$ 
			($n_{jt^{new}k} := 1$, $k_{jt^{new}} := k$, increment $m_{jk}$)
\RETURN updated $\mathbf{s}$
\end{algorithmic}
\end{algorithm}

\begin{algorithm}[h]
\caption{\textsf{removeCustomer}($j, k, \mathbf{s}^{old}$): Removes existing customer eating dish $k$ from restaurant $j$. }
\label{alg:removeCustomer}
\begin{algorithmic}
\STATE $\mathbf{s} := \mathbf{s}^{old}$
\STATE Sample $t_{ji}$ in proportional to $n_{jt_{ji}k}$
\STATE Decrement $n_{jt_{ji}k}$ (the customer at $t_{ji}$-th table is removed)
\IF{$n_{jt_{ji}k}$ becomes zero}
	\STATE Remove the unoccupied table $t_{ji}$ from restaurant $j$, decrement $m_{jk}$
\ENDIF
\RETURN updated $\mathbf{s}$
\end{algorithmic}
\end{algorithm}

\newpage

\section{Step-wise Gibbs sampler}

To manipulate emission probability $F(x_\ttt)$ with a conjugate prior, we introduced a similar notation to HCRP, 
which can be intuitively understood.

\begin{algorithm}[h]
\caption{Step-wise Gibbs sweep for HCRP-HMM} \label{alg:gibbs}
\begin{algorithmic}
\STATE {\bfseries Input:} $y_1, \ldots, y_\ttt$: observed emissions
\STATE \leavevmode \phantom{\bfseries Input:} $x_1, \ldots, x_\ttt$: previously inferred states
\STATE \leavevmode \phantom{\bfseries Input:} $\mathbf{s}^{old}$: set of CRP seating arrangements
\STATE \leavevmode \phantom{\bfseries Input:} $\mathbf{F}^{old}$: set of emission distributions
\FOR{ $\ttt = 1, \ldots, T$, in random order }
  \STATE $\mathbf{s}_{1} = \mathsf{removeCustomer}(x_{\ttt+1}, x_t, \mathbf{s}^{old})$ 
  \STATE $r^{old}_3 = \mathsf{getProb}(x_{\ttt+1}, x_t, \mathbf{s}_{1})$
  \STATE $\mathbf{F}_{0} = \mathsf{removeCustomer}(y_{\ttt}, x_{\ttt}, \mathbf{F}^{old})$
  \STATE $r^{old}_2 = \mathsf{getProb}(y_{\ttt}, x_t, \mathbf{s}_{1})$
  \STATE $\mathbf{s}_{0} = \mathsf{removeCustomer}(x_{\ttt}, x_{\ttt-1}, \mathbf{s}_{1})$
  \STATE $r^{old}_1 = \mathsf{getProb}(x_{\ttt}, x_{\ttt-1}, \mathbf{s}_{0})$
  \STATE Sample $x_\ttt^*$ in proportion to $q(x_t)$ where
  \STATE \quad $q(x_t) \propto \left(\frac{\alpha_0 \gamma}{(\alpha_0 + n_{x_{t-1} \cdot \cdot}) (\gamma + m_{\cdot \cdot})} \delta_{x_{t+1}} + p( x_t | S_{0}^{(x_{t-1})} ) \right)$ 
  \STATE \qquad \qquad $ {} \cdot p( y_t | F_{x_t} ) \cdot p( x_{t+1} | S_{0}^{(x_{t})} )$
  \STATE $r^{*}_1 = \mathsf{getProb}(\x^*_{t}, x_{t-1}, \mathbf{s}_{0})$
  \STATE $\mathbf{s}_{1} = \mathsf{addCustomer}(\x^*_{t}, x_{t-1}, \mathbf{s}_{0})$  
  \STATE $r^{*}_2 = \mathsf{getProb}(y_{t}, x_t, \mathbf{F}_{0})$
  \STATE $\mathbf{F}^{*} = \mathsf{addCustomer}(y_{t}, x_{t}, \mathbf{F}_{0})$
  \STATE $r^{*}_3 = \mathsf{getProb}(x_{t+1}, \x^*_{t}, \mathbf{s}_{1})$
  \STATE $\mathbf{s}^{*} = \mathsf{addCustomer}(x_{t+1}, \x^*_{t}, \mathbf{s}_{1})$  
  \STATE $R := \min \left( 1, 
  		\dfrac{ r^*_1 }{r^{old}_1}
  		\dfrac{ r^*_2 }{r^{old}_2}
  		\dfrac{ r^*_3 }{r^{old}_3}
  		\dfrac{ q( x_t ) }{ q( \x^*_t )}
		\right) $
  \STATE $ \langle x_t, \mathbf{s}, \mathbf{F} \rangle := \begin{cases}
  				~\langle x_\ttt^*, \mathbf{s}^*, \mathbf{F}^* \rangle & \mbox{with probability }R \\
  				~\langle x_t, \mathbf{s}^{old}, \mathbf{F}^{old} \rangle & \mbox{otherwise} \\
  \end{cases} $
\ENDFOR
\end{algorithmic}
\end{algorithm}

\newpage

\section{Blocked Gibbs sampler}
For details on the $\mathsf{ForwardBackwardSampling}$ routine, please refer to the literature \cite{scott2002bayesian}.

\begin{algorithm}[h]
\caption{$\mathsf{removeSeq}( \ttt_0, \ttt_1, \mathbf{\x}, \mathbf{s}^{old}, \mathbf{F}^{old} )$:
remove customers for a part of state sequence $( \x_{\ttt_0}, \ldots, \x_{\ttt_1} )$
}
\begin{algorithmic}
  \STATE $L = \ttt_1 - \ttt_0 - 1$
  \STATE $\mathbf{s}_{L} = \mathsf{removeCustomer}(\x_{\ttt_0+L}, \x_{\ttt_1}, \mathbf{s}^{old})$
  \STATE $r^{old}_L = p(\x_{ji} = k|\mathbf{s})$ 
  \STATE $\mathbf{F}_{L} = \mathbf{F}^{old}$
  \FOR{ $\ell = L-1$  \DOWNTO $0$ }
    \STATE $\mathbf{s}_{\ell} = \mathsf{removeCustomer}(\x_{\ttt_b+\ell}, \x_{\ttt_b+\ell-1}, \mathbf{s}_{\ell+1})$ 
    \STATE $\mathbf{F}_{\ell} = \mathsf{removeCustomer}(y_{\ttt_b+\ell}, \x_{\ttt_b+\ell}, \mathbf{F}_{\ell+1})$ 
	  \STATE $r^{old}_\ell = \mathsf{getProb}(\x_{\ttt_b+\ell}, \x_{\ttt_b+\ell-1}, \mathbf{s}_{\ell})
	  					 \cdot \mathsf{getProb}(y_{\ttt_b+\ell}, \x_{\ttt_b+\ell}, \mathbf{F}_{\ell}) $
  \ENDFOR
  \RETURN $\langle \mathbf{s}_0, \mathbf{F}_0, \prod_{\ell=0}^L r^{old}_\ell \rangle$
\end{algorithmic}
\end{algorithm}

\begin{algorithm}[h]
\caption{$\mathsf{addSeq}( \ttt_0, \ttt_1, \mathbf{\x}, \mathbf{s}_0, \mathbf{F}_0 )$:
add customers for a part of state sequence $( \x_{\ttt_0}, \ldots, \x_{\ttt_1} )$
}
\begin{algorithmic}
  \STATE $L = \ttt_1 - \ttt_0 - 1$
  \FOR{ $\ell = 0$ \TO $L-1$ }
    \STATE $r^{*}_\ell = \mathsf{getProb}(\x_{\ttt_b+\ell}, \x_{\ttt_b+\ell-1}, \mathbf{s}_{\ell})
	  					 \cdot \mathsf{getProb}(y_{\ttt_b+\ell}, \x^*_{\ttt_b+\ell}, \mathbf{F}_{\ell}) $
    \STATE $\mathbf{s}_{\ell+1} = \mathsf{addCustomer}(\x^*_{\ttt_b+\ell}, \x^*_{\ttt_b+\ell-1}, \mathbf{s}_{\ell})$ 
    \STATE $\mathbf{F}_{\ell+1} = \mathsf{addCustomer}(y_{\ttt_b+\ell}, \x^*_{\ttt_b+\ell}, \mathbf{F}_{\ell})$ 
  \ENDFOR
  \STATE $r^{*}_L = \mathsf{getProb}(\x_{\ttt_1}, \x^*_{\ttt_1-1}, \mathbf{s}_{L}^*)$
  \STATE $\mathbf{s}^{*} = \mathsf{addCustomer}(\x_{\ttt_1+L}, \x^*_{\ttt_1-1}, \mathbf{s}_{L}^*)$;  $F^{*} = F_{L}^*$
  \RETURN $\langle \mathbf{s}^*, \mathbf{F}^*_{L}, \prod_{\ell=0}^L r^{*}_\ell \rangle$
\end{algorithmic}
\end{algorithm}

\begin{algorithm}[h]
\caption{Blocked Gibbs sweep for HCRP-HMM} \label{alg:blockgibbs}
\begin{algorithmic}
\STATE {\bfseries Input:} $y_1, \ldots, y_\ttt$: observed emissions
\STATE \leavevmode \phantom{\bfseries Input:} $\mathbf{\x} = \x_1, \ldots, \x_T$: previously inferred states
\STATE \leavevmode \phantom{\bfseries Input:} $\mathbf{s}$: set of CRP seating arrangements
\STATE \leavevmode \phantom{\bfseries Input:} $\mathbf{F}$: set of emission distributions
\STATE \leavevmode \phantom{\bfseries Input:} $B$: number of blocks
\STATE Choose block boundaries $\ttt_1, \ldots, \ttt_{B-1} \in \{2, \ldots, T\} $; $\ttt_0 := 1$, $\ttt_B = T$
\FOR{ $b = 0, \ldots, B-1$, in random order }
  \STATE $\langle \mathbf{s}_0,  \mathbf{F}_0, r^{old} \rangle = \mathsf{removeSeq}( \ttt_b, \ttt_{b+1} - \ttt_b, \mathbf{\x}, \mathbf{s}, \mathbf{F}, 0 )$;
  \STATE $\x_\ttt^* = \x_\ttt$ for all $t < \ttt_b$ or $t \ge t_{b+1}$
  \STATE $(\x_{\ttt_b}^*, \ldots, \x_{\ttt_b+L-1}^*) = {}$
  \STATE \qquad $\mathsf{FBSampler}(\mathbf{\hat \pi}|_{S_{0}}, F_{0}, y_{\ttt_b:\ttt_b+L-1}, \x_{\ttt_b-1}, \x_{\ttt_b+L})$ 
  \STATE Calculate $q^{old} = q(\x_{\ttt_b}, \ldots, \x_{\ttt_b+L-1})$ and $q^{*} = q(\x_{\ttt_b}^*, \ldots, \x_{\ttt_b+L-1}^*)$ 
  \STATE $Q^{old} = \mathrm{CRP}( \gamma, H )$
  \STATE $Q^* = \mathrm{CRP}( \gamma, H )$  
  \IF{ $\x_{t_{b+1}}$ refers to a new state in $\mathbf{s}_{0}$ }
  	  \STATE $Q^{*} := \mathsf{addCustomer}( \x_{t_{b+1}}, Q^{*} )$ 
  	  \STATE $Q^{old} := \mathsf{addCustomer}( \x_{t_{b+1}}, Q^{old} )$
  \ENDIF
  \FOR{ $t = t_{b}$ \TO $t_{b+1}-1$}
  	\IF{ $\x_\ttt$ refers to a new state in $\mathbf{s}_{0}$ }
  	  \STATE $q^{old} := q^{old} * \mathsf{getProb}(\x_\ttt, Q^{old})$ 
  	  \STATE $Q^{old} := \mathsf{addCustomer}( \x_\ttt, Q^{old} )$
  	\ENDIF
  	\IF{ $\x^*_\ttt$ = $k^{new}$ }
	  \STATE sample $s \sim Q^{*}$ ; $\x^*_\ttt := s$
  	  \STATE $q^{*} := q^{*} * \mathsf{getProb}(\x^*_\ttt, Q^{*})$
  	  \STATE $Q^{*} := \mathsf{addCustomer}( \x_\ttt, Q^{*} )$
  	\ENDIF
  \ENDFOR
  \STATE $S_{0}^* = S_{0}$; $F_{0}^* = F_{0}$ 
  \STATE $\langle \mathbf{s}^{*}, \mathbf{F}^{*}, r^* = \mathsf{addSeq}(\ttt_0, L, \mathbf{\x}, \mathbf{s}_0, \mathbf{F}_0 )$
  \STATE $R := \min \biggl( 1, 
		\dfrac{ q^{old} }{ q^*} \cdot r^{old} \cdot r^*
		\biggr) $
  \STATE $ \langle \mathbf{\x}, \mathbf{s}, \mathbf{F} \rangle := \begin{cases}
  				~\langle \mathbf{\x}^*, \mathbf{s}^*, \mathbf{F}^* \rangle & \mbox{with probability }R \\
  				~\langle \mathbf{\x}^{old}, \mathbf{s}^{old}, \mathbf{F}^{old} \rangle & \mbox{otherwise} \\
  \end{cases} $
\ENDFOR
\end{algorithmic}
\end{algorithm}

\onecolumn{

\section{Split-Merge sampler}


%
%

\begin{algorithm}
\caption{Split-Merge Sampler for an HCRP-HMM} \label{alg:splitmerge}
\begin{algorithmic}
\STATE {\bfseries Input:} $y_1, \ldots, y_T$: observed emissions
\STATE \leavevmode \phantom{\bfseries Input:} $\x_1, \ldots, \x_T$: previously inferred states
\STATE \leavevmode \phantom{\bfseries Input:} $\mathbf{s}^{old}$: set of CRP seating arrangements
\STATE \leavevmode \phantom{\bfseries Input:} $\mathbf{F}^{old}$: set of emission distributions

\STATE $R^{thr} \sim \mathrm{Uniform}(0, 1)$
\STATE Choose distinct $t_1, t_2 \in \{1, \ldots, T\}$
\STATE Identify all fragments $(b_i, e_i)$ s.t. 
	for all $t \in (b_i, \ldots, e_i)$, $\x_\ttt \in \{\x_{t_1}, \x_{t_2}\} \land t \notin \{t_1, t_2\}$, and
	not contained in other fragments
\STATE Permute fragments randomly
\STATE Let $U$ be the number of fragments
\STATE $\mathbf{s}_{U+1} = \mathbf{s}^{old}$, $\mathbf{F}_{U+1} = \mathbf{F}^{old}$
\IF{ $\x_{t_1} = \x_{t_2}$ }
	\STATE \COMMENT{Try split move}
	\FOR{ $i=U$ \DOWNTO 1 }
		\STATE $ \langle \mathbf{s}_{i}, \mathbf{F}_{i}, r^{old}_i \rangle 
				= \mathsf{removeSeq}(b_i, e_i, \mathbf{\x}, \mathbf{s}_{i+1}, \mathbf{F}_{i+1}) \!\!$
		\STATE $q^{old}_i = 1$
	\ENDFOR
	\STATE $\x^*_{t_2} =$ new $k$ index
  \ELSE
	\STATE \COMMENT{Try merge move}
	\FOR{ $i=U$ \DOWNTO 1 }
		\STATE $ \langle \mathbf{s}_{i}, \mathbf{F}_{i}, r^{old}_i \rangle 
				= \mathsf{removeSeq}(x_{b_i: e_i}, \mathbf{s}_{i+1}, \mathbf{F}_{i+1}) \!\!$
		\STATE $q^{old}_i = \dfrac{\mathsf{SeqProb}(\mathbf{\hat \pi}|_{\mathbf{s}_{i+1}}, F_{i}, y_{b_i:e_i}, \x_{b_i-1:e_i+1})}%
{\mathsf{ForwardProb}(\mathbf{\hat \pi}|_{\mathbf{s}_{i+1}}, F_{i}, y_{b_i:e_i}, \x_{b_i-1}, \x_{e_i+1}; \{ \x_{t_1}, \x_{t_2} \} ) } $
	\ENDFOR
	\STATE $\x^*_{t_2} = \x_{t_1}$
\ENDIF
	\STATE \COMMENT{Remove customers that accounts for transitions around $x^{old}_{t_2}$}
	\STATE $\mathbf{F}_0 = \mathsf{removeCustomer} (y_{t_2}, \x_{t_2}, \mathbf{F}_1)$
	\STATE $p^{old}_0 = p(y_{t_2} |  \mathbf{F}_{\x_{t_2}} )$
	\STATE $\mathbf{s}_0 := \mathbf{s}_1$
	\IF{$t_2-1$ is not in any fragment }
		\STATE $\mathbf{s}_0 := \mathsf{removeCustomer}(\x_{t_2}, \x_{t_2-1}, \mathbf{s}_0)) $
		\STATE $r^{old}_0 *= \mathsf{getProb}(\x_{t_2}, \x_{t_2-1}, \mathbf{s}_1))$
	\ENDIF
	\IF{$t_2+1$ is not in any fragment }
		\STATE $\mathbf{s}_0 := \mathsf{removeCustomer}(\x_{t_2}+1, \x_{t_2},\mathbf{s}_0)) $
		\STATE $r^{old}_0 *= \mathsf{getProb}(\x_{t_2+1} \x_{t_2},\mathbf{s}_0))$
	\ENDIF
	\STATE $q^{old}_0 = q^*_0 = 1$
	\STATE (continue to Algorithm~\ref{alg:splitmerge2})
\end{algorithmic}
\end{algorithm}

\begin{algorithm}
\caption{Split-Merge Sampler for an HCRP-HMM (continued)} \label{alg:splitmerge2}
\begin{algorithmic}
	\STATE \COMMENT{Add customers that accounts for transitions around $x^{*}_{t_2}$}
	\STATE $ p^{*}_0 = p(y_{t_2} | \mathbf{F}^*_{\x^*_{t_2}} )$
	\STATE $\mathbf{F}^*_1 = \mathsf{addCustomer} (y_{t_2}, \x^*_{t_2}, \mathbf{F}^*_0)$

	\STATE $\mathbf{s}^*_1 := \mathbf{s}_0$
	\IF{$t_2+1$ is not in any fragment }
		\STATE $r^{*}_0 *= getProb(\x_{t_2+1} \x_{t_2},\mathbf{s}^*_1))$
		\STATE $\mathbf{s}^*_1 := \mathsf{addCustomer}(\x_{t_2}+1, \x_{t_2},\mathbf{s}^*_1)) $
	\ENDIF
	\IF{$t_2-1$ is not in any fragment }
		\STATE $r^{*}_0 *= getProb(\x_{t_1} | \x_{t_2-1},\mathbf{s}^*_1))$
		\STATE $\mathbf{s}^*_1 := \mathsf{addCustomer}(\x_{t_2}, \x_{t_2}-1,\mathbf{s}^*_1)) $
	\ENDIF

\IF{ $\x_{ t_1} = \x_{t_2}$ }
	\STATE \COMMENT{Try split move}
	\FOR{ $i = 1$ \TO $U$ }
		\STATE $\x^*_{b_i}, \ldots, \x^*_{b_i+L_i-1} = \mathsf{LimitedFBSampler}( \mathbf{\hat \pi}|_{s^*_{i+1}}, F^*_{i}, y_{b_i:b_i+L_i-1}, \x^*_{b_i-1}, \x^*_{b_i+L}; \{ \x^*_{t_1}, \x^*_{t_2}\} )$
		\STATE $q^{*}_i = \dfrac{\mathsf{SeqProb}(\mathbf{\hat \pi}|_{s^*_{i+1}}, F^*_{i}, y_{b_i:e_i}, \x^*_{b_i-1:e_i+1})}
					 	        {\mathsf{ForwardProb}(\mathbf{\hat \pi}|_{s^*_{i+1}}, F^*_{i}, y_{b_i:e_i}, \x^*_{b_i-1}, \x^*_{e_i+1}; \{ \x^*_{t_1}, \x^*_{t_2} \} ) } $
		\STATE $\langle \mathbf{s}^*_{i+2}, \mathbf{F}^*_{i+1}, r^*_i \rangle 
				= \mathsf{addSeq}(b_i, e_i, \mathbf{\x}^*, \mathbf{s}^*_{i+1}, \mathbf{F}^*_{i}) $
	\ENDFOR
\ELSE
\STATE \COMMENT{Try merge move}
	\FOR{ $i = 1$ \TO $U$ }
		\STATE $ 
		R^{cur} = \prod_{i'=0}^{i-1} \dfrac{r^*_i}{q^*_i} \cdot \prod_{i=0}^I \dfrac{r^{old}_i}{r^{old}_i}$
		\IF{ $R^{thr} \ge R^{cur} $ }
			\STATE rejection determined, exit loop
		\ENDIF
		\STATE $\x^*_{b_i}, \ldots, \x^*_{e_i} = \x_{t_1}$
		\STATE $q^*_i = 1$
		\STATE $\langle \mathbf{s}^*_{i}, \mathbf{F}^*_{i-1}, r^{*}_i \rangle 
				= \mathsf{addSeq}(b_i, e_i, \mathbf{\x}, \mathbf{s}_{i+1}, \mathbf{F}_i) $
	\ENDFOR
\ENDIF
	\STATE $R = \prod_{i=0}^I \dfrac{r^*_i}{r^{old}_i} \cdot \prod_{i=1}^I \dfrac{q^{old}_i}{q^*_i}$
	\STATE $\langle \mathbf{\x}, \mathbf{s}, \mathbf{F} \rangle =
	\begin{cases}
		\langle \mathbf{\x}^{*}, \mathbf{s}^{*}, \mathbf{F}^{*} \rangle & R^{thr} < R \\
		\langle \mathbf{\x}^{old}, \mathbf{s}^{old}, \mathbf{F}^{old} \rangle & \mbox{otherwise}
	\end{cases}$
\end{algorithmic}
\end{algorithm}
}

\end{document}